\documentclass[conference]{IEEEtran}
\IEEEoverridecommandlockouts
\usepackage{cite}
\usepackage{amsmath,amssymb,amsfonts}
\usepackage{graphicx}
\usepackage{textcomp}
\usepackage{multirow}
\usepackage{xcolor}
\usepackage{booktabs}
\usepackage{algorithm}
\usepackage{algpseudocode}
\usepackage{hyperref}
\hypersetup{hidelinks,
	colorlinks=true,
	allcolors=black,
	pdfstartview=Fit,
	breaklinks=true}
\def\BibTeX{{\rm B\kern-.05em{\sc i\kern-.025em b}\kern-.08em
    T\kern-.1667em\lower.7ex\hbox{E}\kern-.125emX}}
\begin{document}

\title{Single-cell Curriculum Learning-based Deep Graph Embedding Clustering
}

\author{\IEEEauthorblockN{Huifa Li}
\IEEEauthorblockA{\textit{Shanghai Key Laboratory of Trustworthy} \\
\textit{Computing} \\
\textit{East China Normal University}\\
Shanghai, China \\
huifali@stu.ecnu.edu.cn}
\and
\IEEEauthorblockN{Jie Fu}
\IEEEauthorblockA{\textit{Department of Computer Science} \\
\textit{Stevens Institute of Technology}\\
Hoboken, USA \\
jfu13@stevens.edu}
\and
\IEEEauthorblockN{Xinpeng Ling}
\IEEEauthorblockA{\textit{Shanghai Key Laboratory of Trustworthy} \\
\textit{Computing} \\
\textit{East China Normal University}\\
Shanghai, China \\
xpling@stu.ecnu.edu.cn}

\and
\IEEEauthorblockN{Zhiyu Sun}
\IEEEauthorblockA{\textit{Shanghai Key Laboratory of Trustworthy} \\
\textit{Computing} \\
\textit{East China Normal University}\\
Shanghai, China \\
zysun@stu.ecnu.edu.cn}
\and
\IEEEauthorblockN{Kuncan Wang}
\IEEEauthorblockA{\textit{Shanghai Key Laboratory of Trustworthy} \\
\textit{Computing} \\
\textit{East China Normal University}\\
Shanghai, China \\
kuncan.wang@stu.ecnu.edu.cn}
\and
\IEEEauthorblockN{Zhili Chen*\thanks{* Corresponding author: Zhili Chen.}}
\IEEEauthorblockA{\textit{Shanghai Key Laboratory of Trustworthy} \\
\textit{Computing} \\
\textit{East China Normal University}\\
Shanghai, China \\
zhlchen@sei.ecnu.edu.cn}
}

\maketitle

\begin{abstract}
The swift advancement of single-cell RNA sequencing (scRNA-seq) technologies enables the investigation of cellular-level tissue heterogeneity. Cell annotation significantly contributes to the extensive downstream analysis of scRNA-seq data. However, The analysis of scRNA-seq for biological inference presents challenges owing to its intricate and indeterminate data distribution, characterized by a substantial volume and a high frequency of dropout events. Furthermore, the quality of training samples varies greatly, and the performance of the popular scRNA-seq data clustering solution GNN could be harmed by two types of low-quality training nodes: 1) nodes on the boundary; 2) nodes that contribute little additional information to the graph. 
To address these problems, we propose a single-cell curriculum learning-based deep graph embedding clustering (scCLG). 
We first propose a Chebyshev graph convolutional autoencoder with multi-criteria (ChebAE) that combines three optimization objectives, including topology reconstruction loss of cell graphs, zero-inflated negative binomial (ZINB) loss, and clustering loss, to learn cell-cell topology representation. 
Meanwhile, we employ a selective training strategy to train GNN based on the features and entropy of nodes and prune the difficult nodes based on the difficulty scores to keep the high-quality graph. 
Empirical results on a variety of gene expression datasets show that our model outperforms state-of-the-art methods.
The code of scCLG will be made publicly available at \href{https://github.com/LFD-byte/scCLG}{https://github.com/LFD-byte/scCLG}.
\end{abstract}


\section{Introduction}
The advent of single-cell RNA sequencing (scRNA-seq) technologies has enabled the measurement of gene expressions in a vast number of individual cells, offering the potential to deliver detailed and high-resolution understandings of the intricate cellular landscape. The analysis of scRNA-seq data plays a pivotal role in biomedical research, including identifying cell types and subtypes, studying developmental processes, investigating disease mechanisms, exploring immunological responses, and supporting drug development and personalized therapy \cite{kharchenko2021triumphs}. Cell annotation is the fundamental step in analyzing scRNA-seq data. In early research, various traditional clustering methods have been applied such as K-means, spectral clustering, hierarchical clustering and density-based clustering. However, scRNA-seq data are so sparse that most of the measurements are zeros. The traditional clustering algorithm often produces suboptimal results.
    
Several clustering methods have been developed to address these limitations. CIDR \cite{lin2017cidr}, MAGIC \cite{dijk2017magic}, and SAVER \cite{huang2018saver} have been developed to initially address the issue of missing values, commonly referred to as dropouts, followed by the clustering of the imputed data. Despite the benefits of imputation, these methods encounter challenges in capturing the intricate inherent structure of scRNA-seq data. Alternative strategies, such as SIMLR \cite{wang2017visualization} and MPSSC \cite{park2018spectral}, utilize multi-kernel spectral clustering to acquire robust similarity measures. Nevertheless, the computational complexity associated with generating the Laplacian matrix hinders their application to large-scale datasets. Additionally, these techniques fail to account for crucial attributes of transcriptional data, including zero inflation and over-dispersion.

In recent years, deep learning has shown excellent performance in the fields of image recognition and processing, speech recognition, recommendation systems, and autonomous driving \cite{lecun1998gradient,hochreiter1997long,kipf2016semi,fu2024differentially,li2024dp}. 
Some deep learning clustering methods have effectively emerged to model the high-dimensional and sparse nature of scRNA-seq data such as scziDesk \cite{chen2020deep}, scDCC \cite{tian2021model}, and scDeepCluster \cite{tian2019clustering}. These models implement auto-encoding architectures. However, they often ignore the cell-cell relationships, which can make the clustering task more challenging. Recently, the emerging graph neural network (GNN) has deconvoluted node relationships in a graph through neighbor information propagation in a deep learning architecture. scGNN \cite{wang2021scgnn} and scGAE \cite{luo2021topology} combine deep autoencoder and graph clustering algorithms to preserve the neighborhood relationships. However, their training strategies largely ignore the importance of different nodes in the graph and how their orders can affect the optimization status, which may result in suboptimal performance of the graph learning models.

In particular, curriculum learning (CL) is an effective training strategy for gradually guiding model learning in tasks with obvious difficulty levels \cite{bengio2009curriculum}. Curriculum learning has applications in natural language processing, computer vision, and other fields that require processing complex data. However, research on scRNA-seq data clustering is still blank, and the impact of traditional curriculum learning methods retaining all data on removing difficult samples on the model has not been explored yet.

\begin{figure*}[htbp]
    \centering
    \includegraphics[width=0.85\linewidth]{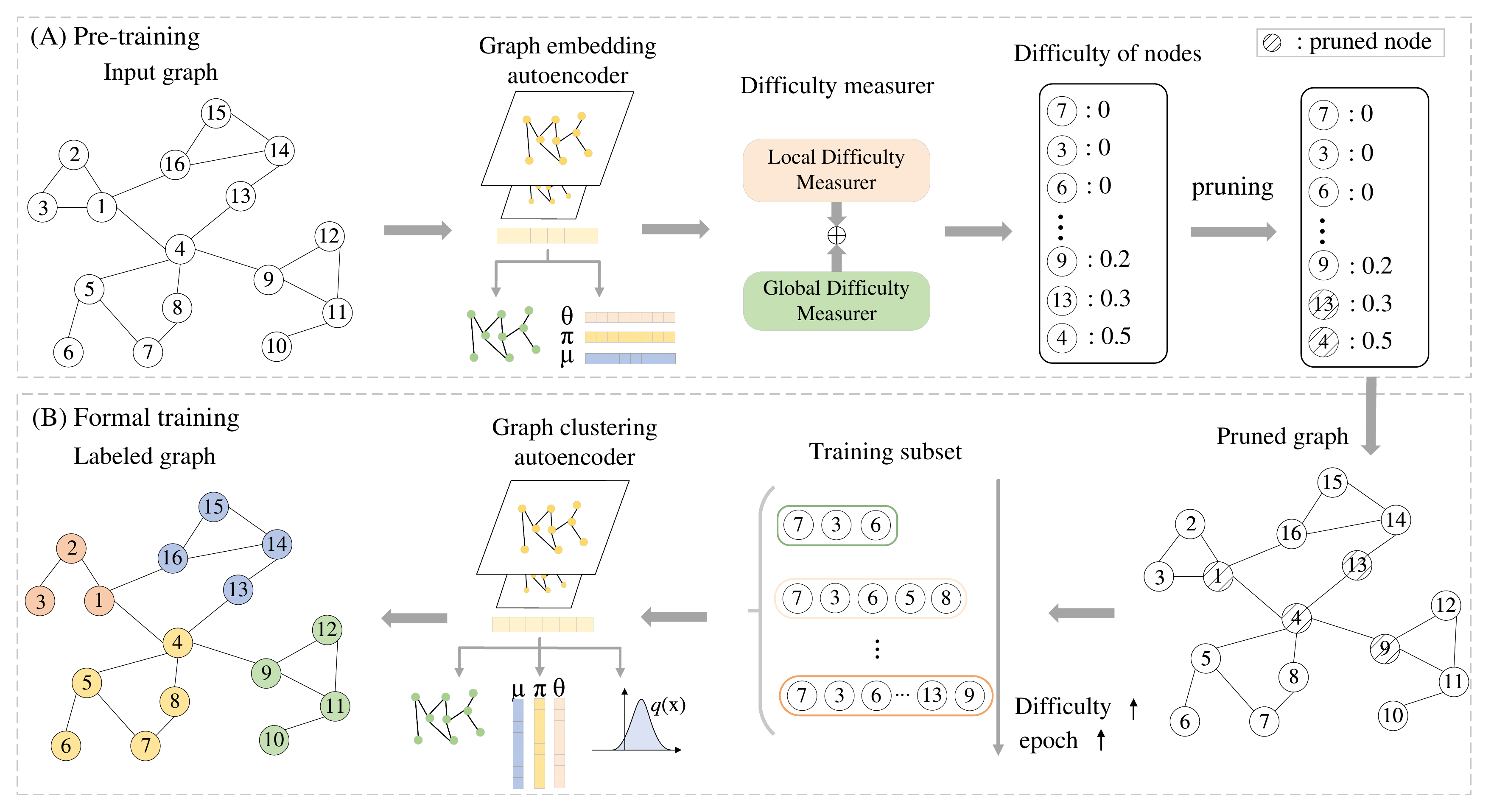}
    \caption{Framework of scCLG. (A) Pre-training: pretraining the proposed ChebAE with adjacency matrix decoder and ZINB decoder. Then calculate node difficulty using a hierarchical difficulty measurer and prune the data. (B) Formal training: using all three criterias to optimize the model in more detail from easy to hard pattern with pruned data.}
    \label{fig:framework}
\end{figure*}

Motivated by the above observations, we propose here a single-cell curriculum learning-based deep graph embedding clustering name scCLG, which simultaneously learns cell-cell topology representations and identifies cell clusters from an autoencoder following an easy-to-hard pattern (Fig. \ref{fig:framework}). 
We first propose a Chebyshev graph convolutional autoencoder with multi-criteria (ChebAE) to preserve the topological structure of the cells in the low-dimensional latent space (Fig. \ref{fig:ae}). 
Then, with the help of feature information, we design a hierarchical difficulty measurer, in which two difficulty measurers from local and global perspectives are proposed to measure the difficulty of training nodes. The local difficulty measurer computes local feature distribution to identify difficult nodes because their neighbors have diverse labels; the global difficulty measurer identifies difficult nodes by calculating the node entropy and graph entropy. 
After that, the most difficult nodes will be pruned to keep the high-quality graph. 
Finally, scCLG can combine three optimization objectives, including topology reconstruction loss of cell graphs, zero-inflated negative binomial (ZINB) loss, and clustering loss, to learn cell-cell topology representation, optimize cell clustering label allocation, and produce superior clustering results.

The main contributions of our work are summarized below:
\begin{itemize}
    \item We propose a single-cell curriculum learning-based deep graph embedding clustering called scCLG, which integrates the meaningful training order into a Chebyshev graph convolutional autoencoder to capture the global probabilistic structure of data.
    \item scCLG constructs a cell graph and uses a Chebyshev graph convolutional autoencoder to collectively preserve the topological structural information and the cell-cell relationships in scRNA-seq data.
    \item To the best of our knowledge, this is the first article to incorporate curriculum learning with data pruning into a graph convolutional autoencoder to model highly sparse and overdispersed scRNA-seq data.
    \item We evaluate our model alongside state-of-the-art competitive methods on 7 real scRNA-seq datasets. The results demonstrate that scCLG outperforms all of the baseline methods.
\end{itemize}

\section{Related Work}

\noindent \textbf{scRNA-seq clustering.}  With the advent of deep learning (DL), more recent works have utilized deep neural networks to automatically extract features from scRNA-seq data for enhancing feature representation. 
scDC \cite{tian2019clustering} simultaneously learns to feature representation and clustering via explicit modeling of scRNA-seq data generation. 
In another work, scziDesk \cite{chen2020deep} combines deep learning with a denoising autoencoder to characterize scRNA-seq data while proposing a soft self-training K-means algorithm to cluster the cell population in the learned latent space. 
scDCC \cite{tian2021model} integrates prior knowledge to loss function with pairwise constraints to scRNA-seq.
The high-order representation and topological relations could be naturally learned by the graph neural network. 
scGNN \cite{wang2021scgnn} introduces a multi-modal autoencoder framework. This framework formulates and aggregates cell–cell relationships with graph neural networks and models heterogeneous gene expression patterns using a left-truncated mixture Gaussian model. 
scGAE \cite{luo2021topology} builds a cell graph and uses a multitask‑oriented graph autoencoder to preserve topological structure information and feature information in scRNA‑seq data simultaneously. However, the above clustering methods overlook the learning difficulty of different samples or nodes.

\noindent \textbf{Curriculum learning.} Curriculum learning, which mimics the human learning process of learning data samples in a meaningful order, aims to enhance the machine learning models by using a designed training curriculum, typically following an easy-to-hard pattern \cite{bengio2009curriculum}. 
The CL framework consists of two components: a difficulty measurer which measures the difficulty of samples and a training scheduler which arranges the ordered samples into training. The key to CL is how to define the promising measurer. SPCL \cite{jiang2015self} takes into account both prior knowledge known before training and the learning progress during training. CLNode \cite{wei2023clnode} measures the difficulty of training nodes based on the label information. SMMCL \cite{gong2019multi} assumes that different unlabeled samples have different difficulty levels for propagation, so it should follow an easy-to-hard sequence with an updated curriculum for label propagation.
scSPaC \cite{zhao2022advanced} utilizes an advanced NMF for scRNA-seq data clustering based on soft self-paced learning, which gradually adds cells from simple to complex to our model until the model converges. However, the above CL methods don't utilize the structural information of nodes in graph neural networks and don't consider the impact of difficult nodes on the graph.

\section{PRELIMINARIES}
In this section, we first introduce some notations, symbols, and necessary background. Then we present the Chebyshev graph convolution.
\subsection{Notations}
Let $\mathcal{G}=(\mathcal{V}, \mathcal{E}, \mathbf{X})$ be an undirected cell graph, where $\mathcal{V}  = \{v_1, v_2,$ $\dots, v_{n_c}\}$ is a set of $n_{c}$ nodes associated with different cells; $e_{ij} \in \mathcal{E}$ specifies the existence of an edge between the $i^{th}$ and $j^{th}$ nodes; and $\mathbf{X}$ is the node feature matrix and $x_{ij}$ element is the count of the $j^{th}$ gene in the $i^{th}$ cell. Let $\mathbf{A} \in \mathbb{R}^{n_c \times n_c}$ be the adjacency matrix of $\mathcal{G}$, where $a_{ij} = 1$ if $v_i$ and $v_j$ are connected, otherwise $a_{ij}$ is set equal to zero. 
The graph Laplacian $\mathbf{L} = \mathbf{D} - \mathbf{A} \in \mathbb{R}^{\mathcal{N} \times \mathcal{N}}$, where $I_{\mathcal{N}}$ is the identity matrix, and $\mathbf{D} \in \mathcal{R}^{\mathcal{N} \times \mathcal{N}}$ is the diagonal degree matrix with $\mathbf{D}_{ii} = \sum_j \mathbf{A}_{ij}$. 
KNN algorithm is employed to construct the cell graph and each node in the graph represents a cell \cite{yu2022zinb}.

\subsection{Chebyshev Graph Convolution}
Chebyshev graph convolution (ChebConv) is a variant of graph convolutional networks that uses Chebyshev polynomials to approximate the feature decomposition of graph Laplacian matrices, thereby achieving convolution operations on graph data. The theoretical foundation of ChebConv is graph signal processing and spectrogram theory, which introduces the concept of graph signal processing into graph convolutional networks. The ChebConv layer is defined as follows:

\begin{align} 
    \mathbf{H} = \sum_{k=1}^K \mathbf{Z}^{(k)} \cdot \Theta^{(k)}
\end{align}

where $K$ represents the order of Chebyshev polynomials used to approximate graph convolution kernels. $\Theta$ is the layer’s trainable parameter and $Z^{(k)}$ is computed recursively by:
\begin{align} 
    \mathbf{Z}^{(1)} &= \mathbf{X} \\
    \mathbf{Z}^{(2)} &= \hat{\mathbf{L}} \cdot \mathbf{X} \\
    \mathbf{Z}^{(k)} &= 2 \cdot \hat{\mathbf{L}} \cdot \mathbf{Z}^{(k-1)} - \mathbf{Z}^{(k-2)}
\end{align}

where $\hat{\mathbf{L}}$ denotes the scaled and normalized Laplacian $\frac{2\mathbf{L}}{\lambda_{max}} - \mathbf{I}$. $\lambda_{max}$ is the largest eigenvalue of $\mathbf{L}$ and $\mathbf{I}$ is the identity matrix.

Compared with basic GCN, ChebConv effectively reduces the model's parameter count and computational complexity by transforming graph convolution operations into approximations of Chebyshev polynomials, while maintaining its ability to capture graph structures.

\section{Proposed Approach}
In this section, 
we firstly present our idea of multi-criteria ChebConv graph autoencoder.
Secondly, we introduce how the scCLG model parameters can be learned using a meaningful sample order. 
Finally, we elaborate the proposed scRNA-seq data clustering algorithm by combining the above two points.

\subsection{Multi-Criteria ChebConv Graph Autoencoder}
\begin{figure}[htbp]
    \centering
    \includegraphics[width=1.00\linewidth]{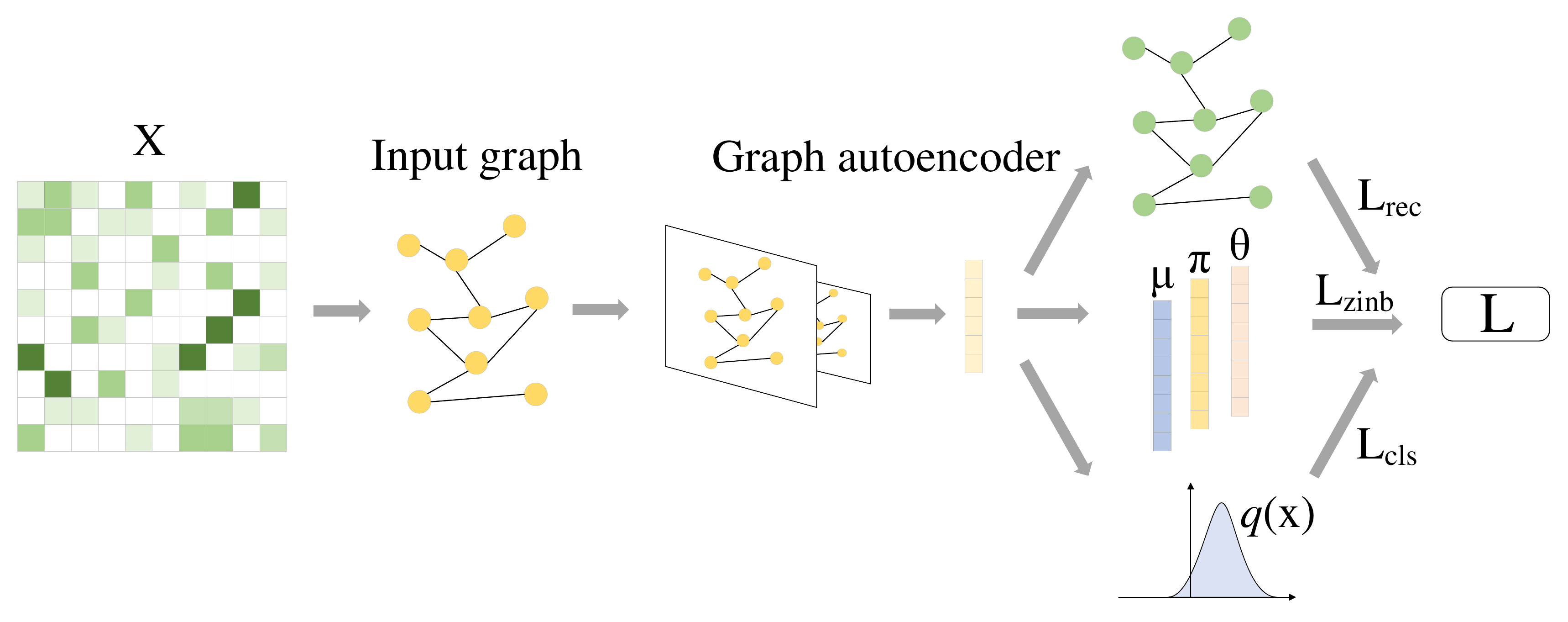}
    \caption{The model architecture of multi-criteria ChebAE. ChebAE integrates three loss components: reconstruction loss, ZINB loss, and a clustering loss to optimize the low-dimensional latent representation.}
    \label{fig:ae}
\end{figure}

As shown in Fig. \ref{fig:ae}, to capture the cell graph structure and node relationships, we developed a variant of the graph convolution autoencoder that uses a stacked topology Chebyshev graph convolutional network as the graph encoder. Compared with basic GCN, ChebConv effectively reduces the model's parameter count and computational complexity by transforming graph convolution operations into approximations of Chebyshev polynomials. 
We use three different criterias to map the encoded compressed vector from different perspectives and jointly optimize the modeling ability of the autoencoder. 
The gene expression matrix $\mathbf{X}$ and normalized adjacency matrix $\mathbf{A}$ are used inputs. 
Through the graph encoder, the feature dimension of each node will be compressed to a smaller size, and the compressed vector features will be decoded by three loss components: reconstruction loss ($L_{rec}$), ZINB loss ($L_{zinb}$), and clustering loss ($L_{cls}$). These criterias share encoder parameters to decompose an optimization objective into three optimization objectives for better capturing the cell-cell relationship:

\begin{align}
    L = L_{rec} + L_{zinb} + L_{cls}
\end{align}

More detailed optimization information about $L_{rec}$, $L_{zinb}$ and $L_{cls}$ is shown below.

\subsubsection{Reconstruction Loss}
Given that the majority of the structure and information inherent in the scRNA-seq data $\mathbf{X}$ is conserved within the latent embedded representation generated by the scCLG encoder.
The adjacency matrix decoder of the graph autoencoder can be defined as the inner product between the latent embedding:
\begin{align}
    \mathbf{Z} &= f_E(\mathbf{X}) \\
    \mathbf{A}_{rec} &= \sigma(\mathbf{Z}^T\mathbf{Z})
\end{align}

where, $f_E$ represents the scCLG encoder function; $\mathbf{A}_{rec}$ is the reconstructed adjacency matrix. Therefore, the reconstruction loss of $\mathbf{A}$ and $\mathbf{A}_{rec}$ should be minimized in the learning process as below:
\begin{align}
    L_{rec} = ||\mathbf{A} - \mathbf{A}_{rec}||_2^2
\end{align}

\subsubsection{ZINB Loss}
In order to more effectively capture the structure of scRNA-seq data by decoding from the latent embedded representation $\mathbf{Z}$, we integrate the ZINB model into a Chebyshev graph convolutional autoencoder to capture the global probability structure of the scRNA-seq data.
Based on this foundation, we propose to apply the ZINB distribution model to simulate the data distribution to capture the characters of scRNA-seq data as follows:
\begin{align}
    \mathrm{NB}(x|\mu,\theta) &= \frac{\Gamma(x+\theta)}{x!\Gamma(\theta)} (\frac{\theta}{\theta+\mu})^{\theta} (\frac{\mu}{\theta+\mu})^{x} \\
    \mathrm{ZINB}(x|\pi,\mu,\theta) &= \pi\delta_0(x) + (1-\pi)\mathrm{NB}(x|\mu,\theta)
\end{align}

where $\mu$ and $\theta$ are the mean and dispersion in the negative binomial distribution, respectively. $\pi$ is the weight of the point mass at zero. The proportion $\frac{\theta}{\theta + \mu}$ replaces the probability p. After that, to model the ZINB distribution, the decoder network has three output layers to compute the three sets of parameters. The estimated parameters can be defined as follows:
\begin{align}
    \hat{\pi} &= \text{sigmoid}(\mathbf{W}_{\pi}f_{D_{zinb}}(\mathbf{Z})) \\
    \hat{\mu} &= exp(\mathbf{W}_{\mu} f_{D_{zinb}}(\mathbf{Z})) \\
    \hat{\theta} &= exp(\mathbf{W}_{\theta}f_{D_{zinb}}(\mathbf{Z}))
\end{align}
where $f_{D_{zinb}}$ is a three-layer fully connected neural network with hidden layers of 128, 256 and 512 nodes. $\mathbf{W}$ represents the learned weights parameter matrices. $\hat{\pi}, \hat{\mu}$ and $\hat{\theta}$ are parameters denoting the estimations of $\pi, \mu$ and $\theta$, respectively. The selection of the activation function depends on the range and definition of the parameters. In terms of the parameter $\pi$, the suitable activation function for it is sigmoid because the interval of $\pi$ is between 0 and 1. Due to the non-negative value of the mean $\mu$ and dispersion $\theta$, we choose the exponential activation function for them. The negative log-likelihood of the ZINB distribution can be used as the reconstruction loss function of the original data $\mathbf{X}$, which can be defined as below:
\begin{align}
    L_{\text{ZINB}} = -\text{log}(\text{ZINB}(\mathbf{X}|\pi,\mu,\theta))
\end{align}
\subsubsection{Clustering Loss}
scRNA-seq clustering clustering as an unsupervised learning task, lacks guidance from labels, which makes it difficult to capture effective optimization signals during the training phase. To overcome this challenge, we apply a clustering module to guide the algorithm to adjust the cluster centers to ensure that the distribution of samples within each cluster is as consistent as possible while minimizing inter-cluster differences. The objective takes the form of Kullback–Leibler (KL) divergence and is formulated as follows:
\begin{align}
    L_{cls} = KL(P||Q) = \sum_i \sum_u p_{iu} \text{log} \frac{p_{iu}}{q_{iu}}
\end{align}
where $q_{iu}$ is the soft label of the embedding node $z_i$ which is defined as the similarity between $z_i$ and cluster centre $\mu_u$ measured by Student’s t-distribution. 
This can be described as follows:
\begin{align}
    q_{ij}=\frac{(1+\|z_{i}-\mu_{j}\|^{2})^{-1}}{\sum_{j^{'}}(1+\|z_{i}-\mu_{j^{'}}\|^{2})^{-1}}
\end{align}
Meanwhile, $p_{iu}$ is the auxiliary target distribution, which puts more emphasis on the similar data points assigned with high confidence on the basis of $q_{iu}$, as below:
\begin{align}
    p_{ij}=\frac{q_{ij}^{2}/\Sigma_{j}q_{ij}}{\sum_{j^{\prime}}(q_{ij^{\prime}}^{2}/\sum_{j^{\prime}}q_{ij^{\prime}})}
\end{align}
Since the target distribution $P$ is defined based on $Q$, the embedding learning of $Q$ is supervised in a self-optimizing way to enable it to be close to the target distribution $P$.

\subsection{Curriculum Learning with Data Pruning}
In this subsection, we first describe the proposed difficulty measurement method from both local and global perspectives and assign a difficulty score to each cell. Based on the difficulty score, we investigate the impact of nodes with higher difficulty on model optimization.

\subsubsection{Hierarchical Difficulty Measurer}
Our Hierarchical Difficulty Measurer consists of two difficulty measures from different perspectives. In this section, we present the definition of two difficulty measures and how to calculate them.

\textbf{Local Difficulty Measurer.}
We introduce how to identify difficult nodes from a local perspective. 
Nodes located at the boundaries of multiple classes may reside in transitional regions within the feature space, leading to less distinct or consistent feature representations, thereby increasing the difficulty of classification. The first type of difficult node should have diverse neighbors that belong to multiple classes. 
Intuitively, features of nodes within the same class tend to be more similar. This is due to the influence of neighboring node features, resulting in nodes with similar connectivity patterns exhibiting comparable feature representations. In order to identify these difficult nodes, we calculate the diversity of the neighborhood's features:
\begin{align}
    D_{local}(u) &= \sum_{v \in \mathcal{N}(u)} S(u, v) \\
    S(u, v) &= \frac{u \cdot v}{||u|| \cdot ||v||}
\end{align}
where $S(u,v)$ denotes the similarity between cell $u$ and cell $v$. A larger $D_{local}(u)$ indicates a more diverse neighborhood. $\mathcal{N}(u)=\{v \in \mathcal{V} |(u,v) \in \mathcal{E} \}$ is the neighborhood of cell $u$. As a result, during neighborhood aggregation, these nodes aggregate neighbors’ features to get an unclear representation, making them difficult for GNNs to learn. By paying less attention to these difficult nodes, scCLG learns more useful information and effectively improves the accuracy of backbone GNNs.

\textbf{Global Difficulty Measurer.}
Then we introduce how to identify difficult nodes from a global perspective. Entropy plays a pivotal role in feature selection as a metric from information theory used to quantify uncertainty. In the process of feature selection, we leverage entropy to assess a feature’s contribution to the target variable. When a feature better distinguishes between different categories of the target variable, its entropy value tends to be relatively low, signifying that it provides more information and reduces overall uncertainty. Consequently, in feature selection, lower entropy values indicate features that offer greater discriminatory power, aiding in the differentiation of target variable categories. We assume nodes that have lower entropy have fewer contributions to the graph. Therefore, this type of node is difficult to classify. Inspired by Entropy Variation \cite{ai2017node}, We consider the node contribution as the variation of network entropy before and after its removal.

For a node $v_i$ in graph $\mathcal{G}$, we define $p(v)$ as probabilities:
\begin{align}
    p(v) &= \frac{D(v)}{\sum_{u \in \mathcal{V}} D(u)}
\end{align}
where $\sum_{v \in \mathcal{V}}p(v)=1$. 

The entropy of the graph is as follows:
\begin{align}
    Ent(\mathcal{G}) &= -\sum_{v \in \mathcal{V}} p(v) \log p(v) \\
    &= -\sum_{v \in \mathcal{V}} \frac{D(v)}{\sum_{u \in \mathcal{V}} D(u)} \log(\frac{D(v)}{\sum_{u \in \mathcal{V}} D(u)}) \\
    &= \text{log}(\sum_{v \in \mathcal{V}} D(v)) -\sum_{v \in \mathcal{V}} \frac{D(v)}{\sum_{u \in \mathcal{V}} D(u)}\log D(v)
\end{align}
where $D(v)$ is the degree of node $v$. $Ent(\mathcal{G})$ is the entropy of graph $\mathcal{G}$ with degree matrix.
    
The global difficulty of the node is as follows:
\begin{align}
    D_{global}(v) &= 1 - \frac{Ent(v)}{\sum_{u \in \mathcal{V}} Ent(u)} \\
     Ent(v) &= Ent(\mathcal{G}) - Ent(\hat{\mathcal{G}}_{v})
\end{align}
where $Ent(v)$ is the change if one node and its connections are removed from the network. $\hat{\mathcal{G}}_{v}$ is the modified graph under the removel of $v$. A lower $Ent(v)$ indicates a lower influence on graph structure and is also more difficult. The global difficulty of node $v$ is to subtract the normalized $Ent(v)$ from 1. 

Considering two difficulty measurers from local and global perspectives, we finally define the difficulty of $v$ as:
\begin{align}
    D(v) = \beta * D_{local} + (1 - \beta) * D_{global} \label{eq:diff_sum}
\end{align}
where $\beta$ is the weight coefficient assigned to each difficulty measurer to control the balance of the total difficulty.

\subsubsection{Data Pruning}
With the hierarchical difficulty measurer, we can get a list of nodes sorted in ascending order of nodes based on difficulty. The node at the end of the list is a nuisance for the overall model learning, so should it be retained? The sources of noise in graph neural networks can be varied, firstly, the attribute information of the nodes may contain noise, which affects the representation of the node features and hence the learning of the GNN. Secondly, the presence of anomalous data may cause the spectral energy of the graph to be "right-shifted", the distribution of spectral energy shifts from low to high frequencies. These noises will not only reduce the performance of the graph neural network but also propagate through the GNN in the topology, affecting the prediction results of the whole network. In order to solve this problem, we designed a data pruning strategy based on the calculated node difficulty. Specifically, we define a data discarding hyperparameter $\alpha$. The value of $\alpha$ is set while balancing data integrity and model generalization performance.
As shown in Fig. \ref{fig:para_prune}, the scRNA-seq clustering performance of the scCLG improves after removing the node features with the highest difficulty which prove our hypothesis.

\subsection{The Proposed scCLG Algorithm}
Our model undergoes a two-phase training process. For the first phase, We pretrain the proposed GNN model ChebAE for discriminative feature learning with an adjacency matrix decoder and ZINB decoder. The number of first phase training rounds is $T_1$ epochs. The output of the encoder is a low dimensional vector which is used to calculate node difficulty using a hierarchical difficulty measurer. We retained the top $1 - \alpha$ of the data with high sample quality for subsequent training. 
For the formal training phase, we use the parameters pretrained and train the model for $T_2$ epochs with pruned data. This phase is the learning of clustering tasks. Unlike the pre-training phase, we use all three criterias to optimize the model in more detail.
We use the pacing function $min(1,2^{\log_2\lambda_0 - \log_2\lambda_0 * \frac{t}{\hat{T}}})$ mentioned in \cite{wei2023clnode} to generate the size of the nodes subset.
We illustrate the detailed information in Algorithm \ref{alg:scclg}.

\begin{algorithm}[htbp]
    \caption{scCLG}
    \label{alg:scclg}
    \begin{algorithmic}[1]
        \Require
        {A scRNA-seq data graph $\mathcal{G}=(\mathcal{V},\mathcal{E},X)$, the GNN model ChebAE, pre-training epochs $T_1$, training epochs $T_2$ and data pruning rate $\alpha$, hyper-parameters $\lambda_0$, $\hat{T}$.}
        \Ensure {The cluster labels $Y$.}
        \State {\textbf{\# Phase 1: pre-training}}
        \State {Initialize parameters of ChebAE}
        \State {Train ChebAE with $Dec_A$ and $Dec_{zinb}$ on $\mathcal{G}$ for $T_1$ epochs}
        \State {Calculate every node difficulty $D(v) \leftarrow \text{Eq}.(\ref{eq:diff_sum})$, $v \in \mathcal{V}$}
        \State {Sort $\mathcal{V}$ according to node difficulty in ascending order}
        \State {Prune ordered nodes at the end with a rate of $\alpha$}
        \State {\textbf{\# Phase 2: formal training}}
        \While {$t < T_2 \ or \ not \ converge$}
            \State {$\beta_t \leftarrow min(1,2^{\log_2\lambda_0 - \log_2\lambda_0 * \frac{t}{\hat{T}}}) \ where \ \beta_t < (1 - \alpha)$}
            \State {Generate training subset $\mathcal{V}_t \leftarrow \mathcal{V}[1,\dots,\lfloor \beta_t \cdot |\mathcal{V}| \rfloor]$}
            \State {Train ChebAE with three criterias on $(\mathcal{V},\mathcal{E},X[\mathcal{V}_t])$}
            \State {$t \leftarrow t + 1$}
        \EndWhile
        \\
    \Return {Predict $Y$ with ChebAE.}
    \end{algorithmic}
\end{algorithm}

\section{Experiments}
\subsection{Setup}
\noindent \textbf{Dataset.} For the former, we collect 7 scRNA-seq datasets from different organisms. The cell numbers range from 870 to 9519, and the cell type numbers vary from 2 to 9.
\begin{table}[htbp]
    \centering
    \caption{Summary of the real scRNA-seq datasets.}
    \label{tab:datasets}
    \begin{tabular}{cccccc}
        \toprule
        \makebox[0.05\textwidth][c]{Dataset} & \makebox[0.05\textwidth][c]{Cells} & \makebox[0.05\textwidth][c]{Genes} & \makebox[0.05\textwidth][c]{Class} & \makebox[0.05\textwidth][c]{Platform} \\
        \midrule
        QS\_Diaphragm & 870 & 23341 & 5 & Smart-seq2 \\
        QS\_Limb\_Muscle & 1090 & 23341 & 6 & Smart-seq2 \\
        QS\_Lung & 1676 & 23341 & 11 & Smart-seq2 \\
        Muraro & 2122 & 19046 & 9 & CEL-seq2 \\
        QS\_Heart & 4365 & 23341 & 8 & Smart-seq2 \\
        Plasschaert & 6977 & 28205 & 8 & inDrop \\
        Wang\_Lung & 9519 & 14561 & 2 & 10x \\
        \bottomrule
    \end{tabular}
\end{table}
    
\noindent \textbf{Baselines.}
The performance of scCLG was compared with two traditional clustering methods (Kmeans and Spectral), and several state-of-the-art scRNA-seq data clustering methods including four single-cell deep embedded clustering methods (scziiDesk, scDC, scDCC and scGMAI) and three single-cell deep graph embedded clustering methods (scTAG, scGAE and scGNN).
    
\begin{itemize}
    \item Deep soft K-means clustering with self-training for single-cell RNA sequence data (\textbf{scziDesk}) \cite{chen2020deep}:
    It combines a denoising autoencoder to characterize scRNA-seq data while proposing a soft self-training K-means algorithm to cluster the cell population in the learned latent space.
    \item Model-based deep embedded clustering method (\textbf{scDC}) \cite{tian2019clustering}: It simultaneously learns to feature representation and clustering via explicit modeling of scRNA-seq data generation.
    \item Model-based deep embedding for constrained clustering analysis of single cell RNA-seq data (\textbf{scDCC}) \cite{tian2021model} It integrates prior information into the modeling process to guide our deep learning model to simultaneously learn meaningful and desired latent representations and clusters.
    \item scGMAI: a Gaussian mixture model for clustering single-cell RNA-Seq data based on deep autoencoder (\textbf{scGMAI}) \cite{yu2021scgmai} It utilizes autoencoder networks to reconstruct gene expression values from scRNA-Seq data and FastICA is used to reduce the dimensions of reconstructed data.
    \item scGNN is a novel graph neural network framework for single-cell RNA-Seq analyses (\textbf{scGNN}) \cite{wang2021scgnn}: It integrates three iterative multi-modal autoencoders and models heterogeneous gene expression patterns using a left-truncated mixture Gaussian model.
    \item A topology-preserving dimensionality reduction method for single-cell RNA-seq data using graph autoencoder (\textbf{scGAE}) \cite{luo2021topology} It builds a cell graph and uses a multitask‑oriented graph autoencoder to preserve topological structure information and feature information in scRNA‑seq data simultaneously.
    \item Zinb-based graph embedding autoencoder for single-cell rna-seq interpretations (\textbf{scTAG}) \cite{yu2022zinb} It simultaneously learns cell–cell topology representations and identifies cell clusters based on deep graph convolutional network integrating the ZINB model.
\end{itemize}
    
\noindent \textbf{Implementation Details.}
In the proposed scCLG method, the cell graph was constructed using the KNN algorithm with the nearest neighbor parameter $k=20$. In the multi-criterias ChebConv graph autoencoder, the hidden fully connected layers in the ZINB decoder are set at 128, 256 and 512. Our algorithm consists of pre-training and formal training, with 1000 and 500 epochs for pre-training and formal training, respectively. Our model was optimized using the Adam optimizer, employing a learning rate of 5e-4 during pre-training and 1e-4 during formal training. The pruning rate $\alpha$ is set to 0.11. For baseline methods, the parameters were set the same as in the original papers.

\begin{table*}[t!]
    \caption{Performance comparison between various baselines on seven real datasets.} 
    \label{tab:cluster_result}
    \centering
    \begin{tabular}{c|c|cccccccccc}
        \toprule
            \makebox[0.05\textwidth][c]{Metric}& \makebox[0.05\textwidth][c]{Dataset} & \makebox[0.05\textwidth][c]{scCLG} & \makebox[0.05\textwidth][c]{scTAG} & \makebox[0.05\textwidth][c]{scGAE} & \makebox[0.05\textwidth][c]{scGNN} & \makebox[0.05\textwidth][c]{scziDesk} & \makebox[0.05\textwidth][c]{scDC} & \makebox[0.05\textwidth][c]{scDCC} & \makebox[0.05\textwidth][c]{scGMAI} & \makebox[0.05\textwidth][c]{Kmeans} & \makebox[0.05\textwidth][c]{Spectral} \\
        \midrule
        \cmidrule{1-12}
        \multirow{7}{*}{ARI}
         & QS\_Diaphragm & \textbf{0.9836} & 0.9628 & 0.5638 & 0.5646 & 0.9517 & 0.6479 & 0.8895 & 0.4111 & 0.9110 & 0.9170 \\
         & QS\_Limb\_Muscle & \textbf{0.9828} & 0.9813 & 0.5419 & 0.6399 & 0.9743 & 0.5384 & 0.3449 & 0.4899 & 0.8922 & 0.9615 \\
         & QS\_Lung & \textbf{0.7946} & 0.6526 & 0.2797 & 0.3631 & 0.7401 & 0.4504 & 0.2908 & 0.4622 & 0.7329 & 0.7559 \\
         & Muraro & \textbf{0.8959} & 0.8878 & 0.6413 & 0.5080 & 0.6784 & 0.6609 & 0.7100 & 0.5132 & 0.8452 & 0.8741 \\
         & QS\_Heart & \textbf{0.9503} & 0.9371 & 0.2497 & 0.5222 & 0.9324 & 0.4673 & 0.2584 & 0.4368 & 0.8376 & 0.8757 \\
         & Plasschaert & \textbf{0.7907} & 0.7697 & 0.3540 & 0.4272 & 0.4867 & 0.4070 & 0.4668 & 0.5711 & 0.7352 & 0.2916 \\
         & Wang\_Lung & \textbf{0.9527} & 0.9004 & 0.1035 & 0.1771 & 0.8975 & 0.2520 & 0.5998 & 0.1325 & 0.7995 & 0.0345 \\
         \cmidrule{1-12}
         \multirow{7}{*}{NMI}
         & QS\_Diaphragm & \textbf{0.9670} & 0.9346 & 0.7351 & 0.7608 & 0.9210 & 0.7807 & 0.8223 & 0.6836 & 0.8846 & 0.8881 \\
         & QS\_Limb\_Muscle & \textbf{0.9682} & 0.9616 & 0.7398 & 0.7726 & 0.9468 & 0.7048 & 0.4624 & 0.7198 & 0.8911 & 0.9389 \\
         & QS\_Lung & \textbf{0.8318} & 0.8038 & 0.6766 & 0.6642 & 0.7543 & 0.6840 & 0.4982 & 0.7312 & 0.7785 & 0.7976 \\
         & Muraro & \textbf{0.8506} & 0.8399 & 0.7619 & 0.6294 & 0.7349 & 0.7549 & 0.8347 & 0.7168 & 0.8194 & 0.8291 \\
         & QS\_Heart & \textbf{0.9064} & 0.8857 & 0.6039 & 0.6540 & 0.8723 & 0.6531 & 0.4242 & 0.6941 & 0.8299 & 0.8454 \\
         & Plasschaert & \textbf{0.7696} & 0.7379 & 0.5563 & 0.5856 & 0.6469 & 0.6122 & 0.5786 & 0.5711 & 0.6915 & 0.5216 \\
         & Wang\_Lung & \textbf{0.8942} & 0.8210 & 0.3150 & 0.3975 & 0.7965 & 0.1511 & 0.5862 & 0.3432 & 0.7167 & 0.0367 \\
       \bottomrule
    \end{tabular}
\end{table*}
    
\subsection{Clustering Result}
Table \ref{tab:cluster_result} shows the clustering performance of our method against multiple state-of-the-art methods, and the values highlighted in bold represent the best results. Obviously, our method outperforms other baseline clustering methods for clustering performance. For the 7 scRNA-seq datasets, scCLG achieves the best NMI and ARI on all datasets. Meanwhile, we can observe that the general deep graph embedded models have no advantage and the clustering performance is not stable. Specifically, scGNN performs poorly on "Wang\_Lung". The main reason is that the information structure preserved by the cell graph alone cannot address the particularities of scRNA-seq data well, and further data order is necessary, which again proves the superiority of scCLG. The performance of the deep clustering method and traditional clustering method exhibits significant fluctuations across different datasets. However, scCLG still has an advantage. This is because the scCLG could effectively learn the key representations of the scRNA-seq data in a meaningful order so that the model can exhibit a smooth learning trajectory. In summary, we can conclude that scCLG performs better than the other methods under two clustering evaluation metrics.

\subsection{Parameter Analysis}
\begin{figure}[htbp]
    \centering
    \includegraphics[width=1.00\linewidth]{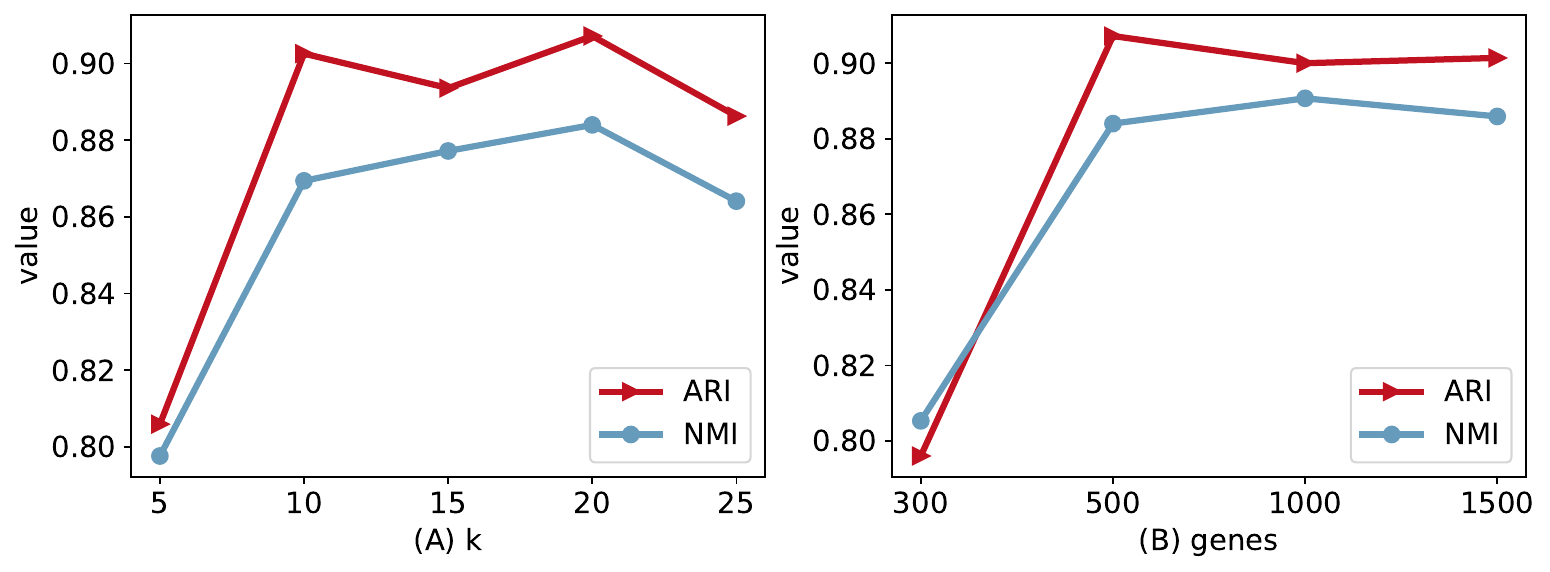}
    \caption{Parameter analysis. (A) Comparison of the average ARI and NMI values with different neighbor parameters $k$. (B) Comparison of the average ARI and NMI values with different numbers of genes.}
    \label{fig:para_gene_k}
\end{figure}
\subsubsection{Different Neighbor Parameter $k$ Analysis}
$k$ represents the number of nearest neighbors to consider when constructing cell graph. 
In order to investigate the impact of $k$, we ran our model with the parameters 5, 10, 15, 25. Fig. \ref{fig:para_gene_k} (A) shows the NMI and ARI values with different numbers of $k$. As depicted in Fig. \ref{fig:para_gene_k} (A), we observe that the two metrics first increase rapidly from parameter 5 to 10, reach the best value at $k =20$, and then decrease slowly from parameter 20 to 25. Therefore, we set the neighbor parameter k as 20 in our scCLG model.

\subsubsection{Different Numbers of Variable Genes Analysis}
In single-cell data analysis, highly variable genes vary significantly among different cells, which helps to reveal the heterogeneity within the cell population and more accurately identify cell subpopulations. 
To explore the impact of the number of selected highly variable genes, we apply scCLG on real datasets with gene numbers from 300 to 1500. Fig. \ref{fig:para_gene_k} (B) shows the line plot of the average NMI and ARI on the 7 datasets selecting 300, 500, 1000 and 1500 genes with high variability, respectively. 
It can be seen that the performance with 500 highly variable genes is better, while the performance with 300 genes is much worse than the others. 
Therefore, to save computational resources and reduce running time, we set the number of selected high-variance genes in the model to 500.

\begin{figure}[htbp]
    \centering
    \includegraphics[width=1.00\linewidth]{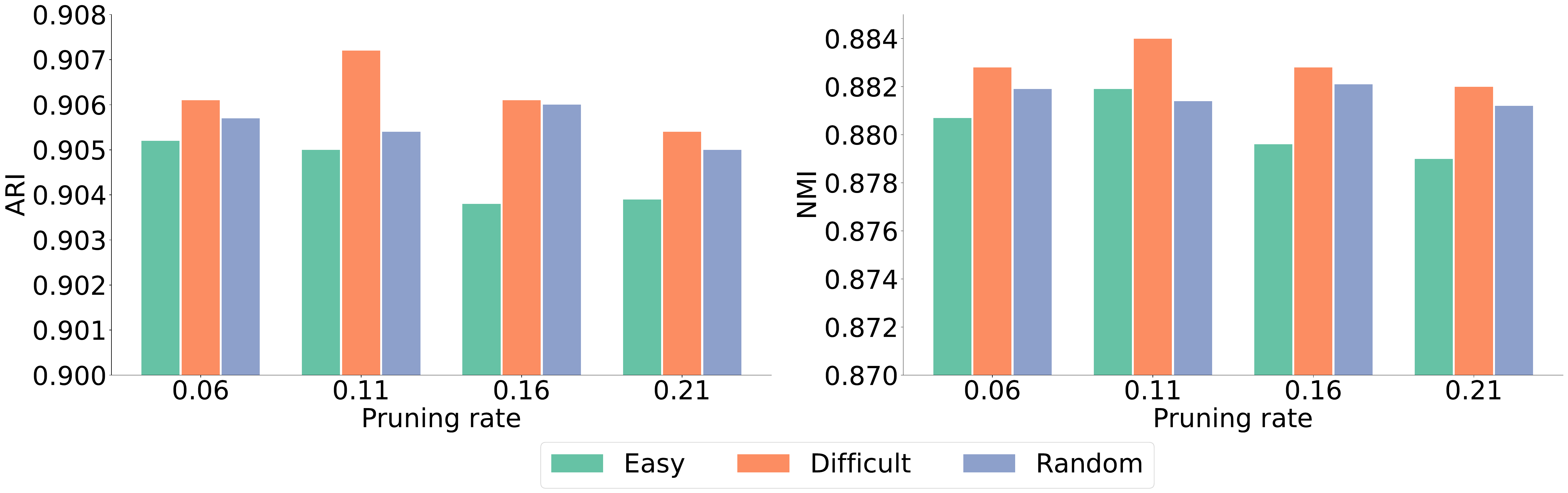}
    \caption{Comparison of the average ARI and NMI values with different data pruning rates and pruning strategies.}
    \label{fig:para_prune}
\end{figure}

\subsubsection{Different Data Pruning Rate Analysis}
In single-cell data analysis, data quality can be improved by pruning lower-quality samples thereby affecting the ability to generalize the model.
To explore the impact of the selected data, we run our model with pruning rate parameters from 0.06 to 0.21 to drop difficult nodes. We also compared our pruning strategy with two different pruning strategies, namely pruning easy nodes and randomly pruning nodes.
Fig. \ref{fig:para_prune} shows the ARI and NMI values with different numbers of $\alpha$ and pruning strategy. 
As depicted in Fig. \ref{fig:para_prune}, we can observe that the best performance is achieved when the $\alpha$ is 0.11 and when difficult nodes are pruned. This indicates that the improvement of data quality can significantly improve the performance of the model. Compared to pruning easy nodes and randomly pruning nodes, pruning difficult nodes brings higher profit because difficult nodes have a negative impact on the representation of the graph. Furthermore, randomly pruning nodes is better than pruning easy nodes, indicating the effectiveness of our difficulty measurer which can assign reasonable difficulty scores to nodes.

\subsection{Ablation Study}
\begin{table}[htbp]
    \caption{Ablation study measured by ARI and NMI values.} 
    \label{tab:ablation}
    \centering
    \begin{tabular}{c|c|cc}
        \toprule
            \makebox[0.05\textwidth][c]{Metric}& \makebox[0.1\textwidth][c]{Methods} & \makebox[0.05\textwidth][c]{scCLG} & \makebox[0.1\textwidth][c]{Without CL} \\
        \midrule
        \cmidrule{1-4}
        \multirow{7}{*}{ARI}
         & QS\_Diaphragm & \textbf{0.9836} & 0.9778 \\
         & QS\_Limb\_Muscle & \textbf{0.9828} & 0.9791 \\
         & QS\_Lung & 0.7946 &  \textbf{0.7947} \\
         & Muraro & \textbf{0.8959} & 0.8897 \\
         & QS\_Heart & 0.9503 & \textbf{0.9530} \\
         & Plasschaert & \textbf{0.7907} & 0.7903 \\
         & Wang\_Lung & \textbf{0.9527} & 0.9527 \\
         \cmidrule{1-4}
         \multirow{7}{*}{NMI}
         & QS\_Diaphragm & \textbf{0.9670} & 0.9579 \\
         & QS\_Limb\_Muscle & \textbf{0.9682} & 0.9613 \\
         & QS\_Lung & 0.8318 & \textbf{0.8321} \\
         & Muraro & \textbf{0.8506} & 0.8468 \\
         & QS\_Heart & 0.9064 & \textbf{0.9088} \\
         & Plasschaert & \textbf{0.7696} & 0.7693 \\
         & Wang\_Lung & \textbf{0.8942} & 0.8942 \\
       \bottomrule
    \end{tabular}
\end{table}
    
In this experiment, we analyzed the effect of each component of the scCLG method. Specifically, we ablated different components in no hierarchical difficulty measurer named Without CL. Table \ref{tab:ablation} tabulates the average ARI and NMI values on the 7 datasets with scCLG. As shown in Table \ref{tab:ablation}, it can be clearly observed that gene screening and extraction of scRNA-seq data from easy to hard patterns improves the clustering performance. For the 7 scRNA-seq datasets, scCLG achieve the best ARI and NMI on 5 of them.
In summary, all components of the scCLG method are reasonable and effective.

\section{Conclusion}

In this research, we propose a single-cell curriculum learning-based deep graph embedding clustering. 
Our approach first utilizes the Chebyshev graph convolutional autoencoder to learn the low-dimensional feature representation which preserves the cell–cell topological structure. Then we define two types of difficult nodes and rank the nodes in the graph based on the measured difficulty to train them in a meaningful manner. Meanwhile, we prune the difficult node to keep the high quality of node features.
Our method shows higher clustering performance against state-of-the-art approaches for scRNA-seq data. Empirical results provide strong evidence that this performance is imputed to the proposed mechanisms and particularly their ability to tackle the difficult nodes.

\bibliographystyle{IEEEtran}
\bibliography{ref}
\end{document}